\documentclass[runningheads]{llncs}
\usepackage{graphicx}
\usepackage{enumitem}
\usepackage{graphicx}
\usepackage{booktabs}
\usepackage{multirow}
\usepackage{diagbox}
\usepackage{subfigure}
\usepackage{bbm}
\usepackage{bm}
\usepackage{color}
\usepackage{cite}

\usepackage{tikz}
\usepackage{comment}
\usepackage{amsmath,amssymb} % define this before the line numbering.
\usepackage{color}
\usepackage[pagebackref,colorlinks]{hyperref}
\usepackage[accsupp]{axessibility}  % Improves PDF readability for those with disabilities.

\begin{document}
% \linenumbers
\pagestyle{headings}
\mainmatter
\def\ECCVSubNumber{100}  % Insert your submission number here

\title{Affective Behaviour Analysis Using Pretrained Model with Facial Prior} % Replace with your title

\begin{comment}
\titlerunning{ECCV-22 submission ID \ECCVSubNumber} 
\authorrunning{ECCV-22 submission ID \ECCVSubNumber} 
\author{Yifan Li$^\star$\inst{1,2} \and
Haomiao Sun$^\star$\inst{1,2} \and
Zhaori Liu\thanks{These authors contribute equally to this work.}\inst{1,2} \and
Hu Han\inst{1,2} \and
Shiguang Shan\inst{1,2} 
\institute{Key Laboratory of Intelligent Information Processing of Chinese Academy of Sciences (CAS), Institute of Computing Technology, CAS, Beijing 100190, China. \and
University of the Chinese Academy of Science, Beijing 100049, China.
\email{liyifan201@mails.ucas.ac.cn}\\
\email{haomiao.sun@vipl.ict.ac.cn}\\
\email{zrliu2000@gmail.com}\\
\email{\{hanhu, sgshan\}@ict.ac.cn}\\
}
}
\institute{Paper ID \ECCVSubNumber}
\end{comment}
%******************

%\begin{comment}
\titlerunning{Affective Behaviour Analysis Using Pretrained Model with Facial Prior}
\author{Yifan Li$^\star$\inst{1,2} \and
Haomiao Sun$^\star$\inst{1,2} \and
Zhaori Liu\thanks{These authors contribute equally to this work.}\inst{1,2} \and
Hu Han\inst{1,2} \and
Shiguang Shan\inst{1,2} 
\institute{Key Laboratory of Intelligent Information Processing of Chinese Academy of Sciences (CAS), Institute of Computing Technology, CAS, Beijing 100190, China. \and
University of the Chinese Academy of Science, Beijing 100049, China.
\email{liyifan201@mails.ucas.ac.cn}\\
\email{haomiao.sun@vipl.ict.ac.cn}\\
\email{zrliu2000@gmail.com}\\
\email{\{hanhu, sgshan\}@ict.ac.cn}\\}}
\authorrunning{Li et al.}
%\end{comment}
%******************

\maketitle

\begin{abstract}
Affective behavior analysis has aroused researchers’ attention due to its broad applications. However, it is labor exhaustive to obtain accurate annotations for massive face images. Thus, we propose to utilize the prior facial information via Masked Auto-Encoder (MAE) pretrained on unlabeled face images. Furthermore, we combine MAE pretrained Vision Transformer (ViT) and AffectNet pretrained CNN to perform multi-task emotion recognition. We notice that expression and action unit (AU) scores are pure and intact features for valence-arousal (VA) regression. As a result, we utilize AffectNet pretrained CNN to extract expression scores concatenating with expression and AU scores from ViT to obtain the final VA features. Moreover, we also propose a co-training framework with two parallel MAE pretrained ViTs for expression recognition tasks. In order to make the two views independent, we randomly mask most patches during the training process. Then, JS divergence is performed to make the predictions of the two views as consistent as possible. The results on ABAW4 show that our methods are effective, and our team reached 2nd place in the multi-task learning (MTL) challenge and 4th place in the learning from synthetic data (LSD) challenge. Code is available \footnote{\url{https://github.com/JackYFL/EMMA_CoTEX_ABAW4}}.
% \dots
\end{abstract}

\section{Introduction}

Affective behaviour analysis such as facial expression recognition (EXPR) \cite{li2020deep}, action unit (AU) recognition \cite{niu2019multi} and valence arousal (VA) regression \cite{nicolaou2011continuous}, raised much attention due to its wide application scenarios. With the superior performance of deep learning, the traditional artificial designed emotional representations are gradually replaced by the deep neural networks (DNNs) extracted ones. While DNNs-based methods perform well in affective behaviour analysis tasks, the limited size of the existing emotion benchmark has hindered the recognition performance and generalization ability. 

Although it costs much to obtain data with accurate emotion annotations, we can acquire unlabeled face images more easily. To capitalize on the massive unlabeled face images, we propose to learn facial prior knowledge using self-supervised learning methods. 
In computer vision field, general self-supervised learning methods can be summarized into two major directions, contrastive learning based methods, e.g., MoCo \cite{he2020momentum}, SimCLR \cite{chen2020simple}, and generative methods, e.g., BeiT \cite{bao2021beit}, Masked Auto-Encoder (MAE) \cite{he2022masked}. SSL-based methods have reached great success, achieving even better performance compared with supervised pretrained ones in some downstream tasks. Inspired by MAE, which randomly masks image patches and utilizes vision transformer \cite{dosovitskiy2020image} (ViT) to reconstruct pixels and learn the intrinsic relationships and discriminant representations of patches, we propose to learn facial prior using MAE on massive unlabeled face images. We suppose MAE could model the relationship of different face components, which contains the prior about face structure and has a better parameter initialization compared with the ImageNet pretrained model. 

Based on MAE pretrained ViT, we propose a simple but effective framework called \underline{E}motion \underline{M}ulti-\underline{M}odel \underline{A}ggregation (EMMA, see Fig. \ref{EMMA}) method for multi-task affective behaviour analysis, i.e., EXPR, AU recognition and VA regression. According to experiment results, we find that it's easier to overfit for VA regression task when finetuning  MAE pretrained ViT for all three tasks. As a result, we propose to use an extra convolutional neural network (CNN) to extract features for VA regression tasks. We found that expression scores could provide pure and intact features for VA regression. Thus we choose to use a CNN (DAN \cite{wen2021distract}) pretrained on an EXPR benchmark AffectNet that contains expression prior knowledge to extract VA features. Furthermore, we also utilize the scores of both EXPR and AU recognition extracted by ViT to aid VA regression. Note that we only finetune the linear layer for VA regression to prevent overfitting.

Moreover, we also propose a masked \underline{Co}-\underline{T}raining method for \underline{EX}pression recognition (masked CoTEX, see Fig. \ref{CoTEX}), in which we use MAE pretrained ViT as the backbones of two views. To form two information-complementary views, we randomly mask most patches of the face images. Inspired by MLCT \cite{xing2018multi} and MLCR \cite{niu2019multi}, we also use Jenson-Shannon (JS) divergence to constrain the predictions from both views to be consistent to the average distribution of two predictions. Different from MLCT or MLCR, we apply JS divergence to single-label classification (e.g., expression) by modifying the entropy, and this method is prepared for supervised learning.

The contributions of this paper can be summarized into the following aspects.

\begin{itemize}[leftmargin=*]
    \item We use face images to pretrain MAE, which shows better performance than ImageNet pretrained MAE for affective behaviour analysis related tasks.
    
    \item We design EMMA, which uses ViT (pretrained on unlabeled face images) and  CNN (pretrained on expression benchmark) to extract features for multi-task affective behaviour analysis. We also propose to use the concatenation of  expression scores extracted by CNN and ViT, and the AU scores extracted by ViT to finetune the linear layer for VA regression.

    \item We propose masked CoTEX for EXPR, which is a co-training framework utilizing JS divergence on two random masked views to make them consistent. We find that a large mask ratio could not only improve training speed and decrease memory usage, but also increase accuracy. Furthermore, we also find that the accuracy could be improved further by increasing the batch size.
    
\end{itemize}

\section{Related Work}
In this section, we first review studies on multi-task affective behaviour analysis and then introduce related work on static image-based EXPR. 

\textbf{Multi-task Affective Behaviour Analysis.} 
The MTL affective behaviour analysis task refers to the simultaneous analysis of EXPR, AU recognition, and VA regression, etc. Unlike the experimental setup in ABAW and ABAW2, where three different tasks were completed independently, ABAW3 presents an integrated metric and evaluates the performance of all three tasks simultaneously. Deng~\cite{deng2022multiple} employs psychological prior knowledge for multi-task estimation, which uses local features for AU recognition and merges the messages of different regions for EXPR and VA. Jeong et al.~\cite{jeong2022multitask} apply the knowledge distillation technique for a better generalization performance and the domain adaptation techniques to improve accuracy in target domains. Savchenko et al.~\cite{kim2022facial} used a lightweight EfficientNet model to develop a real-time framework and improve performance by pre-training based on additional data. Unlike previous methods for MTL, we utilize the facial prior information in both unlabeled face images and expression benchmark to improve the performance.

\textbf{Expression Recognition.} The aim of expression recognition is to recognize basic human expression categories. In the ABAW3 competition, some researchers utilize multi-modal information such as audio and text to improve the model performance and achieve a better ranking~\cite{zhang2022transformer,kim2022facial}. However, these methods have higher requirements for data collection. Therefore, it is worth exploring how to use static images for a more generalized usage scenario. Jeong et al.~\cite{jeong2022facial} use an affinity loss approach, which uses affinity loss to train a feature extractor for images. In addition, they propose a multi-head attention network in a coordinated manner to extract diverse attention for EXPR. Xue et al.~\cite{xue2022coarse} propose a two-stage CFC network that separates negative and positive expressions, and then distinguishes between similar ones. Phan et al.~\cite{phan2022expression} use a pretrained model RegNet~\cite{radosavovic2020designing} as a backbone and introduce the Transformer for better modelling the temporal information. Different from the previous methods, our masked CoTEX uses a masked co-training framework which consists of two ViTs pretrained by MAE on unlabeled face images to achieve better performance.

\section{Proposed Method}
In this section, we first formulate the problem, then introduce EMMA for multi-task affective behaviour analysis and masked CoTEX for EXPR.

\subsection{Formulation}
Let $\mathcal{X}=\{x_i\in \mathbb{R}^{C\times H\times W}, i=1,2,..., N\}$ and $\mathcal{Y}$ denotes face images and according labels, respectively. For multi-task affective behaviour analysis, the labels $\mathcal{Y}=\{y^i_{MTL}, i=1,2,...,N\}$ consists of three sub-task labels, i.e., 
$$y^{i}_{MTL}=\left[ y^{i}_{VA}\in \mathbb{R}^2,y^{i}_{EXP}\in \mathbb{Z},y^{i}_{AU}\in \mathbb{Z}^{12} \right],$$
where $y^{i}_{VA}, y^{i}_{EXP}, y^{i}_{AU}$ indicate VA labels, EXP labels and AU labels, respectively. $y^i_{VA}$ is a two dimension vector representing valence and arousal in the range of $[-1,1]$. $y^i_{EXP}$ is an integer ranging from 0 to 7 representing one of eight expression categories, i.e., neutral, anger, disgust, fear, happiness, sadness, surprise, and other. $y^i_{AU}$ includes 12 AU labels, i.e., AU1, AU2, AU4, AU6, AU7, AU10, AU12, AU15, AU23, AU24, AU25 and AU26. If a face is invisible due to large pose or occlusion, the values of $y^i_{VA}$, $y^i_{EXP}$ and $y^i_{AU}$ can be -5, -1 and 0, respectively.
For EXPR task, $\mathcal{Y}=\{y^i_{EXP}\in \mathbb{Z}, i=1,2,...,N\}$. There are only six expression categories in ABAW4\cite{kollias2022abaw} synthetic expression dataset, i.e., anger, disgust, fear, happiness, sadness, and surprise.
\begin{figure}[t] % [h] forces the figure to be output where it is defined in the code (it suppresses floating)
	\centering
	\includegraphics[width=0.7\columnwidth]{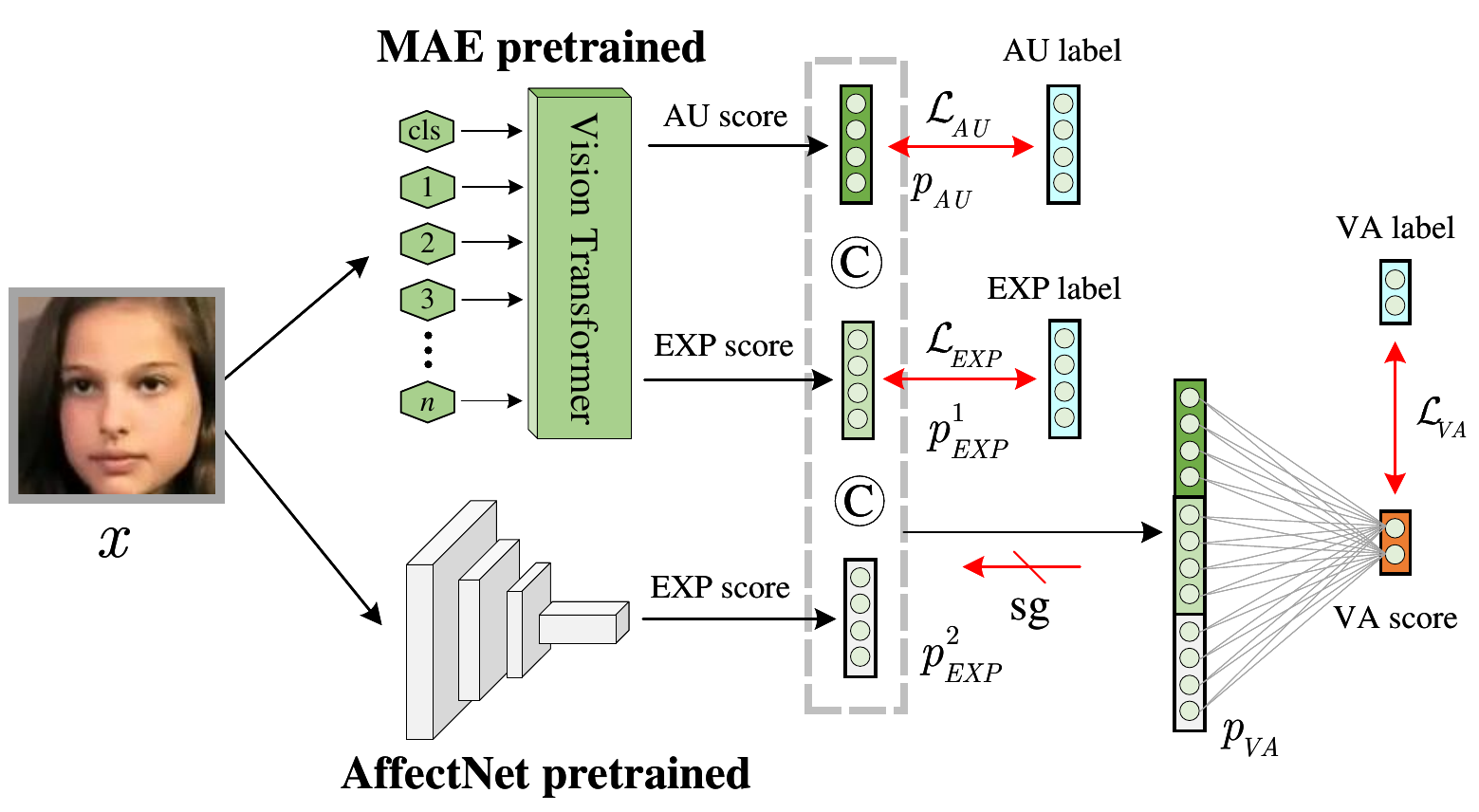} % Example image
% 	\vspace{-10pt}
	\caption{EMMA framework for multi-task affective behaviour analysis.} \label{EMMA}
% 	\vspace{-15pt}
\end{figure} 
\subsection{EMMA}

EMMA shown in Fig. \ref{EMMA} is a two-branch architecture, with one (MAE pretrained ViT) for AU recognition and EXPR tasks, and the other (AffectNet pretrained DAN) for VA recognition task. Assume MAE pretrained ViT and AffectNet pretrained DAN can be denoted as $f_{ViT}$ and $f_{CNN}$, respectively. Given an image $x_i$, we can obtain the AU and EXP prediction score $f_{ViT}(x_i)=\left[ {p_{AU}(x_i),p_{EXP}^1(x_i)}\right]$, and another EXP prediction score  $p_{EXP}^2(x_i) = f_{CNN}(x_i)$. Then the VA feature can be obtained by concatenating three scores.
Finally, we can obtain the VA score $p_{VA}(x_i)$ by passing the VA feature through a two-layer fully connected layer.
Since ViT and CNN are both easy to get overfitting when finetuning the overall network for VA regression task, we stop the gradient of $p_{VA}$ and only finetune the linear layer. The optimization objective $\mathcal{L}$ can be expressed as:
\begin{equation}
    \mathcal{L} = \mathcal{L}_{AU} + \mathcal{L}_{VA} + \mathcal{L}_{EXP},
\end{equation}
where $\mathcal{L}_{AU}$, $\mathcal{L}_{VA}$, $\mathcal{L}_{EXP}$ indicate losses for AU recognition, VA regression, and EXPR recognition, respectively.

For the AU recognition task, the training loss $\mathcal{L}_{AU}$ is the binary cross-entropy loss, which is given by:
\begin{small}
\begin{equation}
    \mathcal{L}_{AU}(p_{AU}(x_i), y_{AU}^i) = - \frac{1}{L}\sum\limits_{j = 0}^{L - 1} \left[{y^{ij}_{AU}\log \sigma(p_{AU}^j({x_i})) + 
     (1 - y^{ij}_{AU})\log (1-\sigma(p_{AU}^j({x_i})))} \right], 
\end{equation}
\end{small}
where L is the number of AUs, and $\sigma$ denotes sigmoid function: $\sigma(x) = \frac{1}{{1 + {e^{ - x}}}}$.

For EXPR task, the training loss $\mathcal{L}_{EXP}$ is the cross-entropy loss:
\begin{equation}
    \mathcal{L}_{EXP} = - \log {\rho^{y^i_{EXP}}},
\end{equation}
where $\rho^{y^i_{EXP}}$ is the softmax probability of the prediction score $p_{EXP}(x_i)$ indexed by the expression label $y^i_{EXP}$.

For VA regression task, we regard Concordance Correlation Coefficient (CCC, see Eq. (\ref{ccc})) loss as training loss $\mathcal{L}_{VA}$:
\begin{equation}
    \mathcal{L}_{VA} = 1 - (CCC^A+CCC^V),
\end{equation}
where $CCC^A$ and $CCC^V$ are the $CCC$ of arousal and valence, respectively.

\begin{figure}[t] % [h] forces the figure to be output where it is defined in the code (it suppresses floating)
	\centering
	\includegraphics[width=0.7\columnwidth]{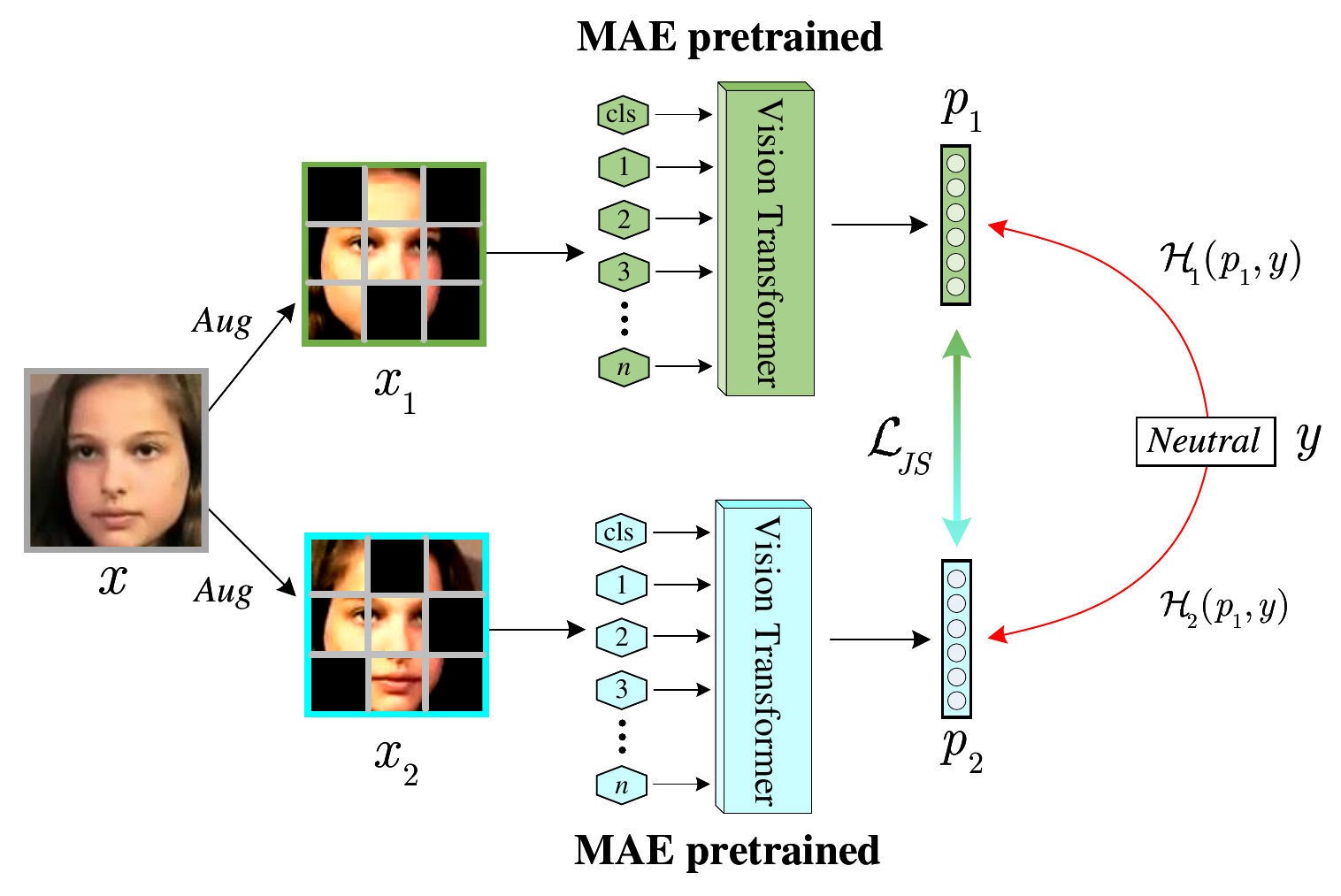} % Example image
% 	\vspace{-10pt}
	\caption{Masked CoTEX framework for EXPR.} \label{CoTEX}
% 	\vspace{-10pt}
\end{figure} 

\subsection{Masked CoTEX}
Masked CoTEX (Fig. \ref{CoTEX}) is designed for EXPR, which is a co-training framework with MAE pretrained ViT in each view. In order to make two views as independent as possible, we randomly mask most patches during training process to form two approximately information-complementary views. Masking random patches in face images means we drop a few patches and put the rest patches into the ViT. Given a face image $x_i$, we can obtain two expression scores $p_1(x_i)$ and $p_2(x_i)$ by two parallel ViTs. JS divergence $\mathcal{L}_{JS}$ is performed to make the predictions of two views consistent. We use the average predictions from two views during inference phase to further improve the performance. The overall optimization objective is:
\begin{equation}
    \mathcal{L} = \lambda \mathcal{L}_{JS} + \mathcal{H}_1 + \mathcal{H}_2,
\end{equation}
$\lambda$ is the hyper-parameter to balance the influence of $\mathcal{L}_{JS}$. $\mathcal{L}$ consists of two components,  JS divergence $\mathcal{L}_{JS}$ and cross-entropy loss $\mathcal{H}_t, t=\{1, 2\}$.

The JS divergence $\mathcal{L}_{JS}$ that constrains the predictions of two views can be expressed as:
\begin{equation}
    {{\cal L}_{JS}}({\rho _1},{\rho _2}) = H(m) - \frac{{H({\rho _1}) + H({\rho _2})}}{2} = \frac{{{D_{KL}}(\left. {{\rho _1}} \right\|m) + {D_{KL}}(\left. {{\rho _2}} \right\|m)}}{2},
\end{equation}
where $\rho_1$ and $\rho_2$ indicate the softmax probabilities of prediction scores, $m$ is the average of $\rho_1$ and $\rho_2$, and $H$ is the entropy of the probability. JS divergence can also be denoted by the average KL divergence which constrains two distributions similar to the average distribution.

In each view, we use cross-entropy loss $\mathcal{H}$ to constrain:
\begin{equation}
    \mathcal{H}_t(\rho_t, y_i) = -log \rho^{y_i}_t, t\in\{1, 2\},
\end{equation}
where $y_i$ is the expression label of face image $x_i$.

\section{Experiment}

\subsection{Benchmarks and Evaluation Metrics}
The benchmark we used is provided by ABAW4 challenge \cite{kollias2022abaw}, which includes two sub-challenges, i.e., the multi-task learning (MTL) challenge and the learning from synthetic data (LSD) challenge. All the face images are aligned and cropped as 112$\times$112.

For MTL challenge, static version of Aff-Wild2 \cite{kollias2022abaw, kollias2021distribution, kollias2021affect, kollias2020deep, kollias2020va, kollias2019expression, kollias2019deep, kollias2018photorealistic, zafeiriou2017aff, kollias2017recognition} are utilized, which contains selected-specific frames from Aff-Wild2. There are in total of around 220K images including training set (142K), validation set (27K) and test set (51K). MTL challenge includes three sub-tasks, i.e., VA regression, EXPR (including 8 categories) and AU recognition (including 12 AUs). The MTL evaluation metrics $P_{MTL}$ consists of three parts:
\begin{equation}
    P_{MTL} = \frac{1}{2}(CC{C_V} + CC{C_A}) + \frac{1}{8}\sum\limits_{i = 0}^7 {F1_{EXPR}^i}  + \frac{1}{{12}}\sum\limits_{i = 0}^{11} {F1_{AU}^i}, 
\end{equation}
where $CCC$ and $F1$ are expressed as:
\begin{equation}\label{ccc}
    CCC(\bm{x},\bm{y}) = \frac{{2\rho {\sigma _{\bm{x}}}{\sigma_{\bm{y}}}}}{{\sigma^2_{\bm{x}} + \sigma^2_{\bm{y}} + {{({\mu_{\bm{x}}} - {\mu_{\bm{y}}})}^2}}},
    \; F1 = 2 \cdot \frac{{precision \cdot recall}}{{precision + recall}},
\end{equation} 
where $\rho$ is the Pearson Coefficient, $\sigma_{\bm{x}}$ is the standard deviation of vector $\bm{x}$, and $\mu_{\bm{x}}$ is the mean value of the vector $\bm{x}$.

For the LSD challenge, the synthetic face images for the EXPR task (in a total of six basic expressions) are generated from some specific frames in the AffWild2. There are around 2770K synthetic face images for training, and 4.6K and 106K real face images in validation and testing set, respectively. The LSD evaluation metrics $P_{LSD}$ can be expressed as:
\begin{equation}
    {P_{LSD}} = \frac{1}{6}\sum\limits_{i = 0}^5 {F1_{LSD}^i}, 
\end{equation}
which is the average of six expression F1 scores.

\subsection{Training Details}
All the data we used are the aligned and cropped version (112$\times$112) provided by ABAW4. We resize the face images to 232$\times$232 (228$\times$228 for LSD task) and perform the random cropping to obtain the image size of 224$\times$224. Furthermore, we also apply random color jitter and horizontal flipping as data augmentation during the training process. We use ViT-base as the backbone for both two methods, which is pretrained by MAE \cite{he2022masked} on face image benchmark CelebA \cite{liu2018large}. We use AffectNet pretrained DAN \cite{wen2021distract} in EMMA. The overall hyper-parameter settings follow common practice of supervised ViT training, which is shown in Table \ref{setting}. 

\setlength{\tabcolsep}{1.1mm}
{ 
  \begin{table*}[t]\scriptsize

    \renewcommand\arraystretch{1.3}
    \centering
    \caption{The F1 scores (in $\%$) comparison for multi-task affective behaviour analysis on ABAW4 validation set. All the results are recorded for the best final score.}    \label{EMMA_comparison}
    \begin{tabular}{lcccc}
    % \small
    % \begin{tabular}{|cl|c|c|c|c|c|c|c|c|c|c|c|c|c|}
      % \specialrule{0em}{0pt}{0pt}
      % \vspace{-0.3cm}
      % \renewcommand\arraystretch{0.8}
      \toprule
      Method  & AU F1    & VA CCC    & EXPR F1    & Final score\\
      \midrule
      Baseline (VGGface, linear evaluation) & - & - & - & 30 \\
        % \midrule
      Clip (RN-50, finetune) & 30.36 & 22.60 & 18.86 & 71.82 \\
        % \midrule
      ResNet50 (ImageNet pretrained, finetune) & 48.85 & 33.55 & 22.00 & 104.40 \\
        % \midrule
      EmotionNet (ABAW3 SOTA, finetune) & 49.53 & 35.41 & 18.27 & 103.21 \\
        % \midrule
      InceptionV3 (ImageNet pretrained, finetune) & 47.53 & 37.85 & 18.66 & 104.46 \\
        % \midrule
      EfficientNet-b2 (AffectNet pretrained, finetune) & 50.60 & 41.74 & 28.19 & 120.53 \\
        \midrule
      ViT-base (MAE pretrained on CelebA, finetune) & 50.41 & 41.21 & \textbf{31.68} & 123.29 \\
        % \midrule
      EMMA (MAE pretrained on CelebA, finetune) & \textbf{50.54} & \textbf{45.88} & 30.28 & \textbf{126.71} \\
      \bottomrule 
    %   \vspace{-15pt}
    \end{tabular}        
\end{table*}}
\setlength{\tabcolsep}{1.3mm}
{ 
  \begin{table*}[t]\scriptsize

    \renewcommand\arraystretch{1.2}
    \centering
    \caption{The F1 score (in \%) and macro accuracy (acc, in \%) for LSD task on ABAW4 validation set.} \label{CoTEX_comparison}
    \begin{tabular}{lcc}
    % \small
    % \begin{tabular}{|cl|c|c|c|c|c|c|c|c|c|c|c|c|c|}
      % \specialrule{0em}{0pt}{0pt}
      % \vspace{-0.3cm}
      % \renewcommand\arraystretch{0.8}
        \toprule
      Method  & Acc    & F1 \\
        \midrule
      Baseline (ResNet-50, ImageNet pretrained) & 67.58 & 59.71 \\
      EfficientNet-b2 (ImageNet pretrained) & 67.71 & 63.33 \\
      EfficientNet-b2 (AffectNet pretrained) & 73.06 & 63.83 \\
      Clip (ViT-base) & 70.84 & 62.13\\
        \midrule
      ViT-base (MAE pretrained on CelebA)& 71.95 & 64.24 \\
     CoTEX (mask ratio 0\%, batch size 64) & 72.84 & 64.67 \\
     Masked CoTEX (mask ratio 75\%, batch size 1024) & \textbf{75.77}  & \textbf{70.68}\\
      \bottomrule 
    %   \vspace{-15pt}
    \end{tabular}        
\end{table*}}

\makeatletter
\def\hlinew#1{
\noalign{\ifnum0=`}\fi\hrule \@height #1 \futurelet
\reserved@a\@xhline}
\makeatother

\setlength{\tabcolsep}{1.3mm}
{ 
  \begin{table*}[t]\scriptsize
    \renewcommand\arraystretch{1.1}
    \centering
    \caption{The training settings for EMMA and Masked CoTEX.} \label{setting}
    \begin{tabular}{l|l|l}
      Config  & EMMA value  & Masked CoTEX value\\
        \hline\hlinew{0.4pt}
      optimizer & AdamW & AdamW \\
      base learning rate & 5e-4 & 5e-4 \\
      weight decay & 0.05 & 0.15 \\
      batch size & 100 & 1024 \\
      clip grad & 0.05 & 0.05 \\
      layer decay & 0.65 & 0.65 \\
      warm up epochs & 5 & 5 \\
      total epochs & 30 & 6 \\
      accumulated iterations & 4 & 4 \\
      drop path rate & 0.1 & 0.1 \\
    \end{tabular}        
\end{table*}}
% \vspace{-15pt}

\setlength{\tabcolsep}{1.3mm}
{ 
  \begin{table*}[t]\scriptsize

    \renewcommand\arraystretch{1.2}
    \caption{The MTL and LSD results of top-5 teams on ABAW4 test set.} \label{CoTEX_test}
    \centering
\subfigure[MTL]{
    \centering
    \begin{tabular}{cccc}
        \toprule
      Team &  Best Overall Score\\
        \midrule
      Situ-RUCAIM3 & 143.61 \\
      \textbf{Ours} &\textbf{119.45} \\
      HSE-NN & 112.99\\
      CNU Sclab & 111.35 \\
      STAR-2022 &  108.55\\
      % HUST-ANT & 107.12 \\
      % SSSIHL-DMACS & 104.06\\
      % DL\_ISIR & 101.87 \\
      % USTC-AC & 93.97 \\
      % CASIA-NLPR & 91.38 \\
      % ITCNU & 68.54 \\
      \bottomrule 
    \end{tabular}}
    \hspace{20pt}
\subfigure[LSD]{\begin{tabular}{cccc}
        \toprule
      Team &  F1\\
        \midrule
      HSE-NN & 37.18\\
      PPAA & 36.51 \\
      IXLAB & 35.87 \\
      \textbf{Ours} & \textbf{34.83} \\
      HUST-ANT & 34.83 \\
      % SZTU-CVGroup & 34.32 \\
      % SSSIHL-DMACS & 33.64 \\
      % STAR-2022 & 32.40 \\
      % USTC-AC & 30.92 \\
      % IMLAB & 30.84 \\
      \bottomrule 
    \end{tabular}}
\end{table*}}

\setlength{\tabcolsep}{1.3mm}
{ 
  \begin{table*}[t]\scriptsize

    \renewcommand\arraystretch{1.2}
    \centering
    \caption{The ablation study for EMMA. All the models are pretrained with MAE.} \label{EMMA_ablation}
    \begin{tabular}{lcccc}
        \toprule
      Method  & AU F1    & VA CCC    & EXPR F1    & Final score\\
        \midrule
      ViT (ImageNet) & 47.67 & 32.61 & 24.65 & 104.92 \\
        % \midrule
      ViT (CelebA) & 50.41 & 41.21 & 31.68 & 123.29 \\
        \midrule
      EMMA (ImageNet) & 45.38 & 42.57 & 27.02 & 114.96 \\
      EMMA (EmotioNet) & 49.00 & 47.11 & 22.51 & 118.62 \\
      EMMA (CelebA) & 50.54 & 45.88 & 30.28 & 126.71 \\
      EMMA (CelebA, $\mathcal{L}_{EXP} (p_{EXP}^1+p_{EXP}^2)$) & 50.56 & 42.90 & 28.74 & 122.20 \\
        \midrule
      EMMA (different epochs ensemble) & 51.56 & 46.59 & 32.29 & 130.44 \\
        % \midrule
      EMMA (different parameters ensemble) & 52.45 & 47.10 & 34.17 & 133.68 \\
      \bottomrule 
    %   \vspace{-15pt}
    \end{tabular}        
\end{table*}}

\setlength{\tabcolsep}{1.3mm}
{ 
  \begin{table*}[t]\scriptsize

    \renewcommand\arraystretch{1.2}
    \centering
    \caption{The ablation study for Masked CoTEX.} \label{CoTEX_ablation}
    \begin{tabular}{lccc}
        \toprule
      Method &  Acc  & F1 \\
        \midrule
      ViT-base (MAE pretrained on CelebA)& 71.95 & 64.24 \\
      CoTEX (mask ratio 0\%, batch size 64) & 72.81 & 64.67 \\
      Masked CoTEX (mask ratio 50\%, batch size 128) & 72.53 & 64.98 \\
      \midrule
      Masked CoTEX (mask ratio 75\%, batch size 64) & 74.17 & 67.67 \\
      Masked CoTEX (mask ratio 75\%, batch size 256) & 74.82 & 68.99 \\
      Masked CoTEX (mask ratio 75\%, batch size 1024) & 75.77 & 70.68 \\
      \midrule
      Masked CoTEX (mask ratio 85\%, batch size 1536) & 78.54 & 73.75 \\
      Masked CoTEX (mask ratio 90\%, batch size 2560) & 81.18 & 78.62 \\
      Masked CoTEX (mask ratio 95\%, batch size 4096) & 83.85 & 82.33 \\
      \bottomrule 
    %   \vspace{-15pt}
    \end{tabular}        
\end{table*}}

\subsection{Recognition Results}
The results for MTL and LSD task on ABAW4 validation set are shown in Table \ref{EMMA_comparison} and Table \ref{CoTEX_comparison}, respectively.

For MTL task, we compare EMMA with baseline (VGG-face) provided by ABAW4, Clip \cite{radford2021learning} (RN-50), ResNet-50 \cite{he2016deep}, EmotionNet \cite{deng2022multiple} (SOTA method in ABAW3 MTL task), InceptionV3 \cite{szegedy2016rethinking}, EfficientNet-b2 \cite{tan2019efficientnet} and ViT-base \cite{dosovitskiy2020image}. We can see from Table \ref{EMMA_comparison} that EMMA achieves the best or competitive performance compared with other methods especially for VA task. EMMA utilizes AffectNet pretrained CNN to extract EXP score, and combines the AU and EXP score from MAE pretrained ViT, which could provide pure and intact features for VA regression. Moreover, we only finetune the linear layer for VA regression which is easy to overfit when finetuning the entire network. Furthermore, the MAE pretrained ViT also contributes to the improvement of the final result, since it could provide more facial structure information which is a better initialization weight for optimization.

For EXPR task, we compare EMMA with baseline (ResNet-50 pretrained  on ImageNet), EfficientNet-b2 \cite{tan2019efficientnet} (pretrained on AffectNet), Clip \cite{radford2021learning} (ViT-base), ViT-base (MAE pretrained on CelebA) and CoTEX. From Table \ref{CoTEX_comparison} we can see that masked CoTEX outperforms all the other methods. It is worth noting that a large mask ratio and batch size is beneficial for improving the performance.

Table \ref{CoTEX_test} shows the best test results of each team in ABAW4. Our team reached 2nd place in the MTL challenge. The Situ-RUCAIM3 team~\cite{zhang2022emotion} outperforms our proposed method due to the utilization of temporal information during the testing phase and more pre-trained models. Meanwhile, our method achieved fourth place out of ten teams in the LSD task. HSE-NN~\cite{savchenko2022hse} uses a model pre-trained with multi-task labels in an external training set, and IXLAB~\cite{jeong2022learning} applies a bagging strategy to obtain better performance.

\subsection{Ablation Study}
We also perform an ablation study on ABAW4 validation set to investigate the effectiveness of each component, which is shown in Table \ref{EMMA_ablation} for EMMA and Table \ref{CoTEX_ablation} for masked CoTEX, respectively.

For EMMA, we can see that face images pretrained MAE has a better performance than ImageNet pretrained one, which indicates that face images may contain facial structure information that is beneficial for affective behaviour analysis. Furthermore, the AffectNet pretrained CNN could provide facial prior knowledge to improve VA regression performance. Moreover, we also explored the influence of face image dataset. We perform experiments on EmotioNet \cite{wang2020tal}, which includes around 1000,000 face images. However, the performance drops when using EmotioNet pretraind model compared with the CelebA pretrained one. We think this may be caused by the quality of face images, such as the resolution, the image noise in face images, etc. We also consider adding $p_{EXP}^2$ for expression task, while the performance has dropped about 3\%, which is probably caused by the noise in expression logits. In the end, we also attempt the ensemble technique to further improve the final performance, and the results show that this technique is useful.

For masked CoTEX, the results show that a large mask ratio contributes to the improvement of the performance. Since a large mask ratio can reduce memory consumption, we can use a larger batch size. We notice that with the increase of the batch size, the performance is improved accordingly. However, since the training set is generated from the validation set, this improvement may not reflect the actual performance on test set. In all our submissions for ABAW4 LSD challenge, CoTEX with 75\% mask ratio ($\lambda$=0, the 4th submission) achieved the highest F1 score which is 4th place in this challenge. But with 95\% mask ratio (the 5th submission), the F1 score is even lower than the one without masking(the 2nd submission). We suppose an excessively large mask ratio may cause overfitting.

\section{Conclusions}
In this paper, we propose two approaches using pretrained models with facial prior, namely EMMA and masked CoTEX, for the ABAW4 MTL and LSD challenges, respectively. We find that the ViT pretrained by MAE using face images performs better on emotion related tasks compared with the ImageNet pretrained one. Moreover, we notice that the expression score is a pure and intact feature for VA regression, which is prone to get overfitting when finetuning the entire network. Furthermore, we propose a co-training framework, in which two views are generated by randomly masking most patches. According to our experiment results, we find that increasing mask ratio and batch size is beneficial to improve the performance on the LSD validation set. In the future, we can also attempt pretraining MAE on different benchmarks.

\clearpage

\bibliographystyle{splncs04}
\bibliography{egbib}

\end{document}